\documentclass[alpha-refs]{wiley-article}
\usepackage{siunitx}
\usepackage{graphicx}
\usepackage{amsmath,amssymb,amsfonts}
\usepackage{algorithmic}
\usepackage{textcomp}
\usepackage{xcolor}
\usepackage{sidecap}
\usepackage{subcaption}
\usepackage{booktabs}
\usepackage[utf8]{inputenc} % to be able to use tildes and ç
\usepackage{cancel}
\usepackage{xspace}
\newcommand{\mtrain}{\mathcal{D}_{train}}
\newcommand{\train}{$\mtrain$\xspace}
\newcommand{\moracle}{\mathcal{D}_{oracle}}
\newcommand{\oracle}{$\moracle$\xspace}
\DeclareMathOperator*{\argmax}{arg\,max}

\usepackage{floatrow}
\newfloatcommand{capbtabbox}{table}[][\FBwidth]

\papertype{Original Article}
\title{O-MedAL: Online Active Deep Learning for Medical Image Analysis}

\author[1]{Asim Smailagic}
\author[3]{Pedro Costa}
\author[1]{Alex Gaudio}
\author[1]{Kartik Khandelwal}
\author[2]{Mostafa Mirshekari}
\author[2]{Jonathon Fagert}
\author[1]{Devesh Walawalkar}
\author[2]{Susu Xu}
\author[3]{Adrian Galdran}
\author[1]{Pei Zhang}
\author[3,4]{Aurélio Campilho}
\author[2]{Hae Young Noh}

\affil[1]{Department of Electrical and Computer Engineering, Carnegie Mellon University, Pittsburgh PA, USA}
\affil[2]{Department of Civil and Environmental Engineering, Carnegie Mellon University, Pittsburgh PA, USA}
\affil[3]{INESC TEC, Porto, Portugal}
\affil[4]{Faculty of Engineering, University of Porto, Portugal}

\corraddress{Asim Smailagic, Department of Electrical and Computer Engineering, Carnegie Mellon University, Pittsburgh PA, USA}
\corremail{asim@cs.cmu.edu}

\fundinginfo{
This work was supported in part by the European Regional Development Fund through the Operational Programme
for Competitiveness and Internationalisation – COMPETE 2020
Programme and in part by the National Funds through the
Funda\c{c}\~{a}o para a Ci\^{e}ncia e a Tecnologia under Project CMUPERI/TIC/0028/2014.
}

\runningauthor{Smailagic et al.}

\begin{document}

\maketitle

\begin{abstract}
Active Learning methods create an optimized labeled training set from unlabeled data.  We introduce a novel Online Active Deep Learning method for Medical Image Analysis.  We extend our MedAL active learning framework to present new results in this paper.  Our novel sampling method queries the unlabeled examples that maximize the average distance to all training set examples.  Our online method enhances performance of its underlying baseline deep network.  These novelties contribute significant performance improvements, including improving the model's underlying deep network accuracy by 6.30\%, using only 25\% of the labeled dataset to achieve baseline accuracy, reducing backpropagated images during training by as much as 67\%, and demonstrating robustness to class imbalance in binary and multi-class tasks.

% Experiments on three medical image datasets show that our method
% requires significantly less labelings, is more accurate, and is more robust to
% class imbalances than existing methods.
% % Our method significantly improves the accuracy of its underlying baseline deep network.
% % Compared to random sampling and uncertainty sampling, the method uses 275 and 200 (out of 768) fewer labeled examples, respectively. 
% For Diabetic Retinopathy detection, we show our method
% improves the baseline model's accuracy by 6.30\%, and the method reaches baseline accuracy when only 25\% of the dataset is
% labeled.

\end{abstract}

% \nocite{costa2018weakly}
% \nocite{smailagicmobilesensing}

\begin{figure}[!t]
\centering
\vspace{2cm}
\includegraphics[width=3.7in]{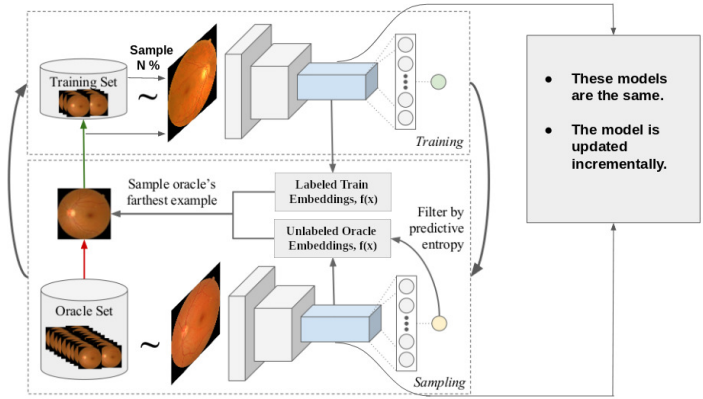}
\caption{\textbf{Proposed Active Learning pipeline.} To solve a supervised classification task, we will use a deep network (DN), an initial labeled dataset \train, an unlabeled dataset \oracle, and an oracle who can label data.  We desire to label as few examples as possible.  
Each active learning iteration, we use the DN to compute a feature embedding for all labeled examples in \train and the top M unlabeled examples from \oracle with highest predictive entropy. We select and label oracle examples furthest in feature space from the centroid of all labeled examples.  
The oracle examples are selected one at a time, and the centroid updated after each labeling.  We train the model on the expanded training set and repeat the process.  In the online setting, the model weights are not reset between iterations and we use only the newly labeled examples and a subset of previously labeled examples.}
%\caption{\textbf{Proposed Active Learning pipeline.}} 
\label{fig_overall}
% \end{SCfigure}
\end{figure}

\section*{\sffamily \Large INTRODUCTION}\label{sec:Introduction}

Active Learning (AL) is an emerging technique for machine learning that aims to reduce the amount of labeled training data necessary for the learning task.  AL techniques are sequential in nature, employing various sampling methods to select examples from an unlabeled  set.  The selected examples are labeled and then used to train the model.  A carefully designed sampling method can reduce the overall number of labeled data points required to train the model and make the model robust to class imbalances \cite{Ertekin:2007:LBA:1321440.1321461} or implicit bias \cite{Richards_2011} in the dataset.

AL assumes  the training process requires labeled data, and secondly that data is costly to label. Medical image analysis is particularly well framed by these assumptions, as the domain offers many opportunities for machine learning solutions, and labeling medical images requires extensive investment of time and effort by trained medical personnel.  In particular, AL can be especially useful in the context of deep learning for medical image analysis, where deep networks typically require large labeled training datasets~\citep{litjens2017survey}.

We introduce MedAL in our prior work. MedAL is a novel AL approach to optimize the selection of unlabeled samples by combining predictive entropy based uncertainty sampling with a distance function on a learned feature space~\citep{smailagic2018medal}.  MedAL's active sampling mechanism minimizes the required labels by selecting only those unlabeled images that are most informative to the model as it trains.

However, MedAL does not address improving computational efficiency.  Each time new labeled examples are added to the training set, MedAL resets the model weights and re-trains the model using all available labeled data. As a result, the method processes training examples many more times than necessary, increasing the time between each sampling step. In a real world application, the trained medical personnel labeling the data would need to wait for the model to finish training, reducing the interactivity and applicability of the system.

To improve the computational performance of MedAL, we introduce ``Online'' MedAL~(O-MedAL) which trains the model incrementally by using only the new set of labeled data and a subset of previously labeled data. By minimizing the training data used in each AL iteration, O-MedAL is computationally faster and more accurate than the original baseline model while retaining all the benefits of MedAL. MedAL is experimentally validated on three medical image diagnosis tasks and modalities: diabetic retinopathy detection from retinal fundus images, breast cancer grading from histopathological images, and melanoma diagnosis from skin images. O-MedAL is compared to MedAL on the retinal fundus image dataset for diabetic retinopathy detection.
To the best of our knowledge, online active learning has never been directly applied to a medical image analysis setting, and it has never been applied to a deep learning setting.  In fact, there is little prior work on online active learning.  We discuss existing approaches and their drawbacks in the Related Work section.

Our \textbf{main contributions} are:
\begin{itemize}
\item \textit{Novelty:} we present a novel AL sampling method that queries the unlabeled examples maximally distant to the centroid of all training set examples in a learned feature space.
\item \textit{Accurate:} we introduce an online training technique that is more accurate than the original baseline model trained on a fully labeled dataset.
\item \textit{Data Efficient:} our method achieves better results with fewer labeled examples than competing methods.
\item \textit{Computationally Efficient:} online training significantly reduces training time and the required number of backpropagated examples.  We also reduce the sampling method runtime from quadratic to linear.
\item \textit{Robust:} our active learning method performs well on binary and multi-class classification problems as well as balanced and unbalanced datasets.
\end{itemize}

The remainder of the paper is structured as follows: in the Related Work section, we discuss relevant prior work on active and online learning approaches. In the Proposed Method section, we describe the sampling process and online training technique.  In the Experiments section, we present results comparing MedAL to common active learning approaches on three medical image datasets.  We also present results comparing O-MedAL to MedAL.  Finally, we discuss the implications of the results and future directions, and we provide conclusions summarizing our work. 

% \section{Related Work}\label{sec:Relatedwork}
\section*{\sffamily \Large RELATED WORK}\label{sec:Relatedwork}

Most AL systems approach the challenge of labeling data by selecting unlabeled examples likely to improve predictive performance.  The AL scenario assumes that an oracle can assign a label to the selected examples. In practice, the process of labeling data can be difficult, time-consuming and expensive. Therefore, it is valuable to design sampling methods that identify the unlabeled examples most informative to the learning task while querying as few labels as possible.

AL was applied to histopathological image analysis on unbalanced data \citep{homeyer_comparison_2012}, cell nucleus segmentation \citep{wen_comparison_2018}, CT scan and MRI analysis \citep{pace_interactive_2015}, and computer-aided diagnosis of masses in digital mammograms \citep{zhao_minimization_2018}.
An AL uncertainty sampling method was created for skin lesion segmentation  \citep{gorriz2017cost}.  Additionally, AL was used in a Multiple Instance Learning framework for tuberculosis detection of chest radiographies \citep{melendez_combining_2016}.

The main AL scenarios include membership query synthesis, pool-based active learning and stream-based active learning. In membership query synthesis, the learner generates an unlabeled input and queries its label~\citep{angluin2004queries}. In pool-based active learning, a pool of unlabeled instances is ranked and the top $k$ subset is labeled~\citep{settles2008curious}. Stream-based active learning assumes a stream of unlabeled data and the learner decides on-the-fly whether to label it ~\citep{dasgupta2008general}.

The sample/query mechanisms mainly include  Query By Committee (QBC)~\citep{freund1997selective,freund1993information}, Expected Error Reduction (EER)~\citep{cohn1995active,melville2004diverse}, Expected Model Change (EMC)~\citep{cai2013maximizing,freytag2014selecting}, and Uncertainty Sampling (US)~\citep{yang2015multi,lewis1994heterogeneous,lewis1994sequential}.

Our work mainly expands on pool-based active learning and US.  US is a sampling technique to select and label the unlabeled examples the model is most uncertain about.  Uncertainty can by computed by predictive entropy, or the entropy of a prediction $p(\hat x | x)$ given an unlabeled input example $x$.  US selects unlabeled examples near the classification boundary without making use of labels.
Building on the US technique, we propose to sample unlabeled examples maximally dissimilar to the training set examples.  % in order to introduce novelty in the training process.

The work of \citep{zhang2018similarity} proposed a pool-based active learning method for highly imbalanced data.  The method automatically pseudo-labels similar examples that have high confidence, and it asks an expert to label examples most similar to rare classes and dissimilar to examples from the common class.  This work complements ours nicely, with key differences.  We sample only the most distant examples across all classes, and we incorporate predictive entropy of the classifier model into the active sampling method.  While \citep{zhang2018similarity} trains two distinct deep networks, we train just one deep network, which is more efficient.  To start the active learning process, their method uses a large initial dataset in which an expert has already labeled many images.  In contrast, our method performs well with an initial dataset of just one labeled example.  We additionally introduce an online learning sampling method to gain further improvements in performance and efficiency, whereas their work focuses on only the active sampling part.

Regarding online active learning, some prior works exist. The work of \citep{murugesan2017active} proposes a multi-task learning algorithm to minimize labeling requests by attempting to infer a label with high confidence from other tasks rather than ask the oracle for the label.   The work of \citep{Sculley2007OnlineAL} addresses a setting where the algorithm requests a label immediately after an example has been classified; the method reduces the computational cost simply because it requires less labels.
The work of \citep{Baram2003OnlineCO} proposes an multi-arm bandits method where learners are selected from the ensemble to sample the next example to be labeled.  The reward function for the bandits problem is classification entropy maximization.  In some cases, the approach has better performance than other ensemble methods evaluated.

More recently, \citep{zhou2017fine} propose an online AL method for biomedical image analysis.
This work presents an active and incremental learning method for medical image data.  While classical AL methods retrain the entire model, in this work a deep convolutional network is continuously fine-tuned at each AL iteration.  We generalize and further improve the online incremental training approach by sampling a random subset of the entire labeled dataset at each AL iteration.
Our results and analysis show that using a random subset of the labeled data is more efficient and achieves better overall performance than using the full dataset for continuous fine tuning.

% \section{Proposed Method}\label{sec:method}
\newcommand{\x}{\mathbf{x}}
\newcommand{\rank}[1]{\phi_\text{rank}\left\langle #1 \right\rangle}
\newcommand{\amr}[1]{ \argmax_{\x\in C_M} \rank{#1} }
\newcommand{\sbr}[1]{\left[#1\right]}
\newcommand{\rbr}[1]{\left(#1\right)}

\section*{\sffamily \Large PROPOSED METHOD}\label{sec:method}

AL techniques developed for classical machine learning methods are not ideal
for \textit{Deep Neural Network} (DNN) architectures.  In particular,
\citep{zhang2016understanding} shows DNNs are capable of fitting a random
labeling of the training data with low uncertainty.
These findings suggest that traditional AL sampling methods based on predictive
uncertainty will be less effective with DNNs.

We present a method specifically tailored to DNNs.  We first describe our novel
AL sampling method and then we show how to train the system using online
learning.

\subsection*{\sffamily \large Sampling Based on Distance between Feature Embeddings}

As shown in Figure \ref{fig_overall}, let \train be the initial training set of labeled examples and \oracle be the unlabeled dataset.  We train a model using \train and then use the model to find a fixed number of the most informative unlabeled examples, $\x^\ast \in \moracle$. These examples are then labeled and added to \train and the model is trained on the new data.  Oracle examples are iteratively labeled in this fashion until the oracle set is exhausted or sufficient performance is attained.

Uncertainty-based methods for choosing $\x^\ast \in \moracle$ evaluate about how informative a given $\x \in \moracle$ is by computing the uncertainty in the model's prediction.
% Since the model is a classifier, its outputs are usually constrained in a well known range (\textit{i.e.} between $0$ and $1$ in a binary classification problem), which makes it possible to directly evaluate the uncertainty of the model on $x$. 
In this work, we depart from the standard practice of only using the model's prediction to choose $\x^\ast$.
Instead, we propose to use a feature embedding function $f(\x)\in \mathbb{R}^n$ in conjunction with uncertainty.

In particular, we first compute the predictive entropy of each unlabeled example in \oracle and then we select only the top $M$ highest entropy examples as a set $\mathcal{C}_M$.  Using this subset of \oracle reduces the number of examples considered and ensures the active learning process ultimately selects unlabeled examples closest to the decision boundary.
\begin{align}
\mathcal{C}_M = \{ \x | \x \in \moracle, \ \  \text{is\_topM}(\text{entropy}(\text{model}(\x)))\}.
\end{align}

To further evaluate the amount of new information an unlabeled example, $\x \in \mathcal{C}_M$, can add to the training set, we evaluate each example's average distance in feature space to each labeled example $\x_i \in \mtrain$ by means of a feature embedding function $f$ and a distance function $d$.
\begin{equation}
s(\x) = \frac{1}{N} \sum^N_{i=1} d(f(\x_i), f(\x)),
\label{eq:feature}
\end{equation}
where $\x_i \in \mtrain$,  $\x \in \mathcal{C}_M$, $f(\x)$ is the feature embedding function and $d(a, b)$ is a Euclidean distance function.

We calculate $s(\x)$ for every example in $C_M \subset \moracle$, then ask the oracle to label the example that maximizes $s$ and add the example to \train.  We repeat this process to select a fixed number of examples from $C_M$.
\begin{align}
\x^\ast = \displaystyle  \argmax_{\x \in \mathcal{C}_{M}} s(\x).
\label{eq:sample}
\end{align}

Finally, we train the model on the new data and repeat the sampling process over again.

The original MedAL paper \citep{smailagic2018medal} evaluated a variety of distance functions and found empirically that Euclidean distance and cosine distance yielded the highest and second highest entropy, respectively, when classifying a random subset of unlabeled examples.  While this evidence suggests either metric is suitable choice for $d$, we choose Euclidean distance.

The following proof shows that using Euclidean distance reduces the computation of $s(\x) \forall \x \in C_M$ from a pairwise distance matrix between examples in \oracle and \train (which is $O(MN)$ operations) to a simple distance of each unlabeled point to the centroid of labeled examples (which is $O(M)$ operations), thereby gaining a quadratic to linear speedup in each AL iteration.  This speedup makes O-MedAL tractable for larger datasets, particularly if there are already many labeled examples (such as in later AL iterations).  A drawback still remains that the method depends on computing feature embeddings, $f(\x)$, for each unlabeled example, which can be expensive.  We also do not know if the linear speedup is possible with cosine distance.

\begin{theorem} In any AL iteration, the unlabeled example $\x^* \in C_M$ can be found with $O(M)$ evaluations of the Euclidean distance function $d$.
  \label{euclidean_proof}
 \end{theorem}

First, note that the embeddings of examples $\x \in C_M$ and examples $\x_i \in \mtrain$ are points in high dimensional space $f(\x) \in \mathbb{R}^d$.  If we assume the space is Euclidean, then the embeddings of examples in \train have a centroid defined by the average of each coordinate: $\mathbf{y} = [y^{(1)}, \dots, y^{(d)}] = [\frac{1}{N}\sum_{i=1}^N f(\x_i)^{(1)}, \dots, \frac{1}{N}\sum_{i=1}^N f(\x_i)^{(d)}]$, where superscript $\ ^{(j)}$ denotes the $j^\text{th}$ coordinate.

Second, note that the scalar score $s(\x)$ for any given example $\x \in C_M$ can be found exactly by evaluating $d(f(\x), f(\x_i))$ for all $\x_i \in \mtrain$.  The score permits an ordering of examples in $C_M$, and therefore we can find $\x^* = \argmax_{\x\in C_M} s(\x)$.  However, evaluating each score exactly is unnecessary. We can use only the knowledge of the ordering to find the maximum; we do not need the specific scores $s(\x)$.  To this end, we introduce an operator $\rank{.}$ which returns the index of the given element in an array sorted from smallest to largest.  Clearly, $\x^* = \argmax_{\x\in C_M} s(\x) = \argmax_{\x\in C_M} \rank{s(\x)}$

We now prove that the ordering of examples $\x \in C_M$ is preserved when $s(\x) = d(f(\x), \mathbf{y})$, where $d$ is Euclidean distance and $\mathbf{y}$ is the centroid of embeddings for elements in \train.

\begin{align}
  \rank{s(\x)} &= \rank{d(f(\x), \mathbf{y})}\\
  \rank{ \frac{1}{N}\sum_{i=1}^N \sqrt{\sum_j^d \rbr{f(\x_i)^{(j)} - f(\x)^{(j)}}^2}}
              &= \rank{ \sqrt{\sum_j^d\rbr{ f(\x)^{(j)} - \frac{1}{N}\sum_{i=1}^N f(\x_i)^{(j)} }^2 }} \label{proof1}
\end{align}

In one dimension, an ordering is preserved through any uniform stretching or shrinking of a number line.  We define the following properties of an ordered variable $z\in \mathbb{R}$ and scalar constant $c$: $\rank{z} = \rank{z+c} = \rank{cz}$.  Furthermore, if $z \geq 0$, then $\rank{z} = \rank{\sqrt{z}} \implies \rank{\sum \sqrt{z}} = \rank{\sum z}$.  We can use these properties to simplify both sides of Eq. \eqref{proof1} by considering $z=f(\x)^{(j)}$ and then removing the square root and constant terms.

\newcommand{\del}[1]{\textcolor{gray}{\cancel{#1}}}
\begin{align}
  \rank{ \sum_{i=1}^N \sum_j^d \rbr{f(\x_i)^{(j)} - f(\x)^{(j)}}^2}
    &= \rank{ \sum_j^d \rbr{f(\x)^{(j)} - \frac{1}{N}\sum_{i=1}^N f(\x_i)^{(j)} }^2} \\
  \rank{ \sum_{i=1}^N \sum_j^d \sbr{ \rbr{f(\x_i)^{(j)}}^2 + \rbr{f(\x)^{(j)}}^2 - 2 f(\x)f(\x_i) }}
    &= \rank{{\frac{1}{N^2}} \sum_j^d  \sbr{Nf(\x)^{(j)}- \sum_{i=1}^N f(\x_i)^{(j)}}^2}\\
  \rank{ \sum_{i=1}^N \sum_j^d \sbr{ \rbr{f(\x)^{(j)}}^2 - 2 f(\x)f(\x_i) }}
    &=\rank{\sum_j^d {N} \sbr{ N \rbr{f(\x)^{(j)}}^2  - 2 f(\x)^{(j)} \sum_{i=1}^N f(\x_i)^{(j)}} + {\sum_j^d\rbr{\sum_{i=1}^N f(\x_i)^{(j)}}^2}}\\
    &=\rank{\sum_j^d \sum_{i=1}^N \sbr{\rbr{f(\x)^{(j)}}^2 - 2 f(\x)^{(j)} f(\x_i)^{(j)} }}.
\end{align}

The right and left sides are equivalent, which proves the ordering is preserved when using the distance to the centroid of the labeled examples.  It follows that $\x^* = \amr{d(f(\x), \mathbf{y})}$ is a valid score function and can be computed with $O(M)$ distance function evaluations.

In following sections, we will describe in detail the feature embedding function $f(\x)$, and specifically how we train the model with new labels.

\subsection*{\sffamily \large Deep Representations as Feature Embeddings}

We train a Convolutional Neural Network (CNN) based architecture to extract
powerful representations from the data and simultaneously solve the image
classification problem.  When interpreted in metric space, the feature
embeddings learned by CNNs are known to encode semantic meaning.  In
particular, nearby elements tend to be visually similar
\citep{mikolov2013efficient,costa2018end}, and conversely, elements that are
far away tend to be visually different.  We therefore make the assumption that
images with mostly similar features will have a smaller distance in embedding
space, while images with mostly different features will have a larger distance.
The assumption that the feature embedding can be interpreted in metric space
is the basis for Eq. (\ref{eq:feature}).

As shown in Figure \ref{fig_overall}, we define the feature embedding
function, $f(\x) \in \mathbb{R}^n$, as the activations of a particular CNN layer.
Since $f(\x)$ is tied to the model, it evolves during training.
As the model performance improves, the model will extract better
representations, leading to better classification accuracy as well as more
informative examples sampled from \oracle.  If $f(\x)$ was statically defined,
the function could be computed off-line and the model reduces to a simple
predictive entropy based sampling method.

\subsection*{\sffamily \large Online Active Learning}

The online learning technique describes how we train the model given a stream of newly labeled examples generated by the AL sampling method.  Online MedAL (O-MedAL) introduces a few changes to MedAL.  First, while MedAL re-initializes the model parameters each AL iteration, O-MedAL maintains the parameters at each new AL iteration and incrementally updates the model.  This change enables significant computational savings because while MedAL re-trains the model once per AL iteration, O-MedAL trains just one model across all AL iterations, and the total number of epochs used to train the model can be an order of magnitude smaller. Second, MedAL trains the model using the full training set \train, while O-MedAL trains on the newly labeled items and a random subset of previously labeled items.  We have found empirically that including a non-null subset of previously labeled data is necessary, and we discuss further in the Discussion section.  Using a subset of available training data also reduces the overall number of examples used to train the model.  Both of these changes result in computational efficiency.  The first change reduces the total number of epochs required to train the model while the second change reduces the number of examples per epoch.

The number of examples that will be used to update model weights during backpropagation increases exponentially, and can be expressed using the equation:

%%% explanation for the equation, text used in experiments:
% The models train for a number of epochs in each AL iteration.  Within each epoch, the model sees a training image once.  Therefore, if a model trains for one AL iteration containing 10 epochs, it processes each training image 10 times.  As we add newly labeled items each AL iteration, the number of examples processed grows exponentially according to the equation:

\begin{align}
    N_t = \sum_{i=1}^t E_i[\ell + p|\mtrain|_{i-1}]
    \qquad\text{ where }\quad|\mtrain|_{i-1} = |\mtrain|_0 + (i-1)\ell, \label{eq:O-MedAL_num_processed}
\end{align}
where $N_t$ is the cumulative number of example images used for backpropagation after $t$ AL iterations, $\ell$ is the number of images labeled each AL iteration, $p \in [0,1]$ is the hyperparameter determining the fraction of previously labeled dataset to use, $|\mtrain|_{i-1}$ is the number of labeled examples in the previous AL iteration and $E_i$ is the number of epochs at the $i^{\text{th}}$ AL iteration.  $|\mtrain|_0$ is the initial number of labeled examples; for O-MedAL, $|\mtrain|_0 = 1$.  The equation is useful to estimate the number of images processed by backpropagation.  It can also be used to estimate how many times an O-MedAL model's parameters have been updated after any given AL iteration; in this case, the cumulative number of weight updates will be $N_t / b$, where $b$ is the minibatch size.

% \section{Experiments}\label{sec:experiments}

\section*{\sffamily \Large EXPERIMENTS}\label{sec:experiments}

We perform experiments on three medical datasets to validate the accuracy and
robustness of our proposed approach.  First, we introduce the datasets. Next, we
describe MedAL and online MedAL implementation details.  Then, we evaluate our
sampling method on the three datasets.  Last, we evaluate our online training
approach.

\subsection*{\sffamily \large Dataset Description}\label{sec:dataset_description}

\begin{figure*}[ht]
\centering
\includegraphics[width=0.85\textwidth]{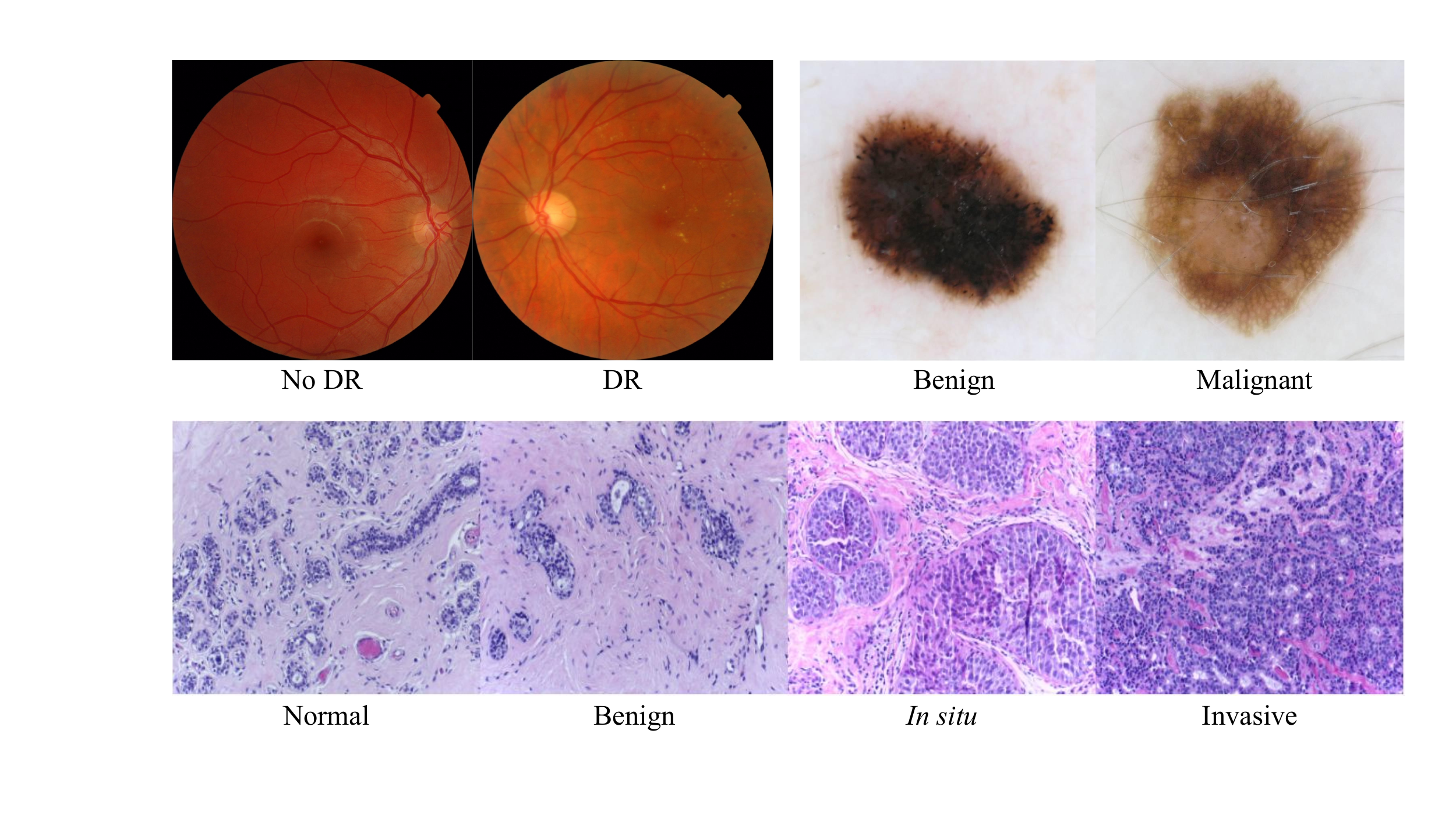}
\caption{\textbf{Examples from all three datasets evaluated in this work.} From left to right, top to bottom: Messidor, Skin Cancer and Breast Cancer datasets. The class labels are shown bellow each of the images.}
\label{fig: datasets}
\end{figure*}

Active learning reduces the amount of labeled data necessary for training, and it is therefore well positioned medical image analysis.  We evaluate MedAL on
three publicly available datasets.  Figure \ref{fig: datasets} presents
representative images in these datasets.

\textbf{Messidor Dataset}
contains 1200 eye fundus images from 654 diabetic and 546 healthy patients
collected from three hospitals in France, we believe between 2005 and 2006.
This dataset was labeled for Diabetic Retinopathy (DR) grading and for Risk of
Macular Edema.  We use Messidor to classify eye fundus images as
healthy (DR grade $= 0$) or diseased (DR grade $> 0$).

\textbf{Breast Cancer Diagnosis Dataset}, originally presented in the ICIAR
2018 Grand Challenge \citep{aresta2018bach}, consists of 400 high resolution
images of breast tissue cells evenly split into four distinct classes: Normal,
Benign, in-situ carcinoma and invasive carcinoma (100 images per class).

\textbf{Skin Cancer dataset} contains 900 cell tissue images classified as
either benign or malignant \citep{gutman2016skin}.  The dataset is
unbalanced, with $80\%$ benign examples and $20\%$ positive examples.  MedAL makes possible the following class balancing technique: in each AL iteration, we
query additional malignant training examples by making use of randomized preprocessing techniques.

\subsection*{\sffamily \large MedAL Implementation}

The underlying convolutional neural network architecture is Inception V3
\citep{szegedy2016rethinking}, with weights pre-trained on ImageNet.  We
replace the top layer with Global Average Pooling and then a Fully-Connected
layer, where the number of hidden units corresponds to the number of output
classes in the dataset.  We use Adam \citep{kingma2014adam} optimizer with
learning rate of $2e-4$ and we use the default recommended values for $\beta_1$
($0.9$) and $\beta_2$ ($0.999$).

At the start of each AL iteration, we reset the model's weights the initial
starting values (after pre-training on ImageNet).  The top layer weights, since
they were not pre-trained on ImageNet, are randomly initialized using the
Glorot method \citep{glorot2010understanding}.

Each AL iteration, we train the model until it reaches $100\%$ training
accuracy.  We perform hyperparameter selection using only the Messidor
validation set and we apply these same hyperparameters to the Skin Cancer and
Breast Cancer datasets in order to avoid using a labeled validation set for
these datasets.  The ability to avoid using a validation set is important to show both
robustness of our method and to show that our method
minimizes the number of labeled images used.  Table \ref{tab: models} shows
the dataset size and the specific hyperparameters that were different across the datasets.

We follow the ORB initialization method described in \citep{smailagic2018medal} to create 
the initial training set \train.
For data pre-processing, we resize all images to $512 \times 512$ pixels and
use the following dataset augmentation: 15 degree random rotation, random
scaling in the range [0.9, 1], and random horizontal flips.

To configure the AL sampler, we use the Euclidean distance function, and we
obtain feature embeddings from the Mixed5 layer of Inception V3.  These choices
are a result of our prior empirical evidence that this combination of layer and
distance function achieves the highest entropy ~\citep{smailagic2018medal}.

\begin{table}
\centering
\caption{\textbf{MedAL Implementation Details.} Showing dataset size and values of parameters used.}
 \begin{tabular}{|c|c|c|c|c|} 
 \hline
 Parameters %& \multicolumn{3}{|c|}{Datasets}\\\cline{2-4} 
                           & Messidor & Breast Cancer & Skin Cancer\\
 \hline
 (train + oracle) size     & 768      & 320           & 700\\
 \hline
 validation set size     & 240 & 0 & 0 \\
 \hline
 test set size     & 192 & 80 & 200 \\
 \hline
 $|\mtrain|_0$ - Initial training set size & 100      & 30            & 100\\
 \hline
 $\ell$ - Imgs labeled each AL iter   & 20       & 5             & 10\\
 \hline
 $M$ - Num max entropy samples & 50       & 30            & 50\\
 \hline
 \end{tabular}
\label{tab: models}
\end{table}

\subsection*{\sffamily \large Online Active Learning (O-MedAL) Implementation}\label{O-MedAL_implementation}

We start with the configuration defined in the MedAL Implementation Details section, and then make the following changes. 

First, we replace the Inception V3 network with a ResNet18 network \citep{resnet} pre-trained on ImageNet.  ResNet18 performs nearly as well as Inception V3 on the fully labeled dataset, and ResNet18 is a smaller model.  Since one of the reasons to consider O-MedAL is computational efficiency, we decided to use ResNet18.  We use the SGD optimizer with a Nesterov momentum of 0.9, learning rate of 0.003, weight decay of 0.01.  The batch size is 48.  We replace the last layer with a linear layer followed by Sigmoid.  The feature embeddings are extracted from the output of the "layer 2" layer.  We consider ResNet18 converged on Messidor at 80 epochs.  ResNet18 with these hyperparameters is the network used for all O-MedAL evaluation experiments.

Second, regarding Messidor, we use an 80/20 random split stratified across hospitals.  The split is re-computed each time a model is trained.  
It assigns 949 images to the combined train and oracle sets, and the remaining 238 images to the test set. For these tests, we also correct the published errata associated with Messidor's dataset (this includes removal of 13 duplicate images from one of the hospitals).

We use the same pre-processing transformations used for MedAL evaluation, and we also use the same hyperparameters for Messidor; the number of max entropy samples is $M=50$ and images added per iteration is $\ell=20$.  For results described in the O-MedAL Evaluation section, we do not use ORB initialization for MedAL or O-MedAL models, and more specifically, $|\mtrain|_0 = 1$ randomly sampled and labeled image.

Third, we introduce a hyperparameter, $p \in [0.0, 1.0]$, that determines at any given AL iteration what percent of previously labeled examples in \train should be used to train the model.  The same images are used for all epochs in a given AL iteration.  This parameter is the independent variable of our O-MedAL experiments.  It was previously described in Equation \ref{eq:O-MedAL_num_processed}. 

Fourth, since the method is online, we do not reset the weights of the model at the start of each AL iteration.

Fifth, for the O-MedAL evaluation, O-MedAL and MedAL models identified by a patience parameter are trained for at most 150 epochs per AL iteration, with an early stopping patience of 5, 10 or 20.  O-MedAL models that do not specify patience are trained for a fixed 10 epochs per AL iteration.

Last, we rewrite the code in a new implementation, migrating from Keras to PyTorch.  The new code is available online at \url{https://www.github.com/adgaudio/O-MedAL}.

\subsection*{\sffamily \large Distance-based Sampling Method Evaluation}\label{sec:acquisition_eval}

\begin{figure}[t]
\centering
\begin{subfigure}[t]{0.45\linewidth}
\centering
\includegraphics[width=\textwidth]{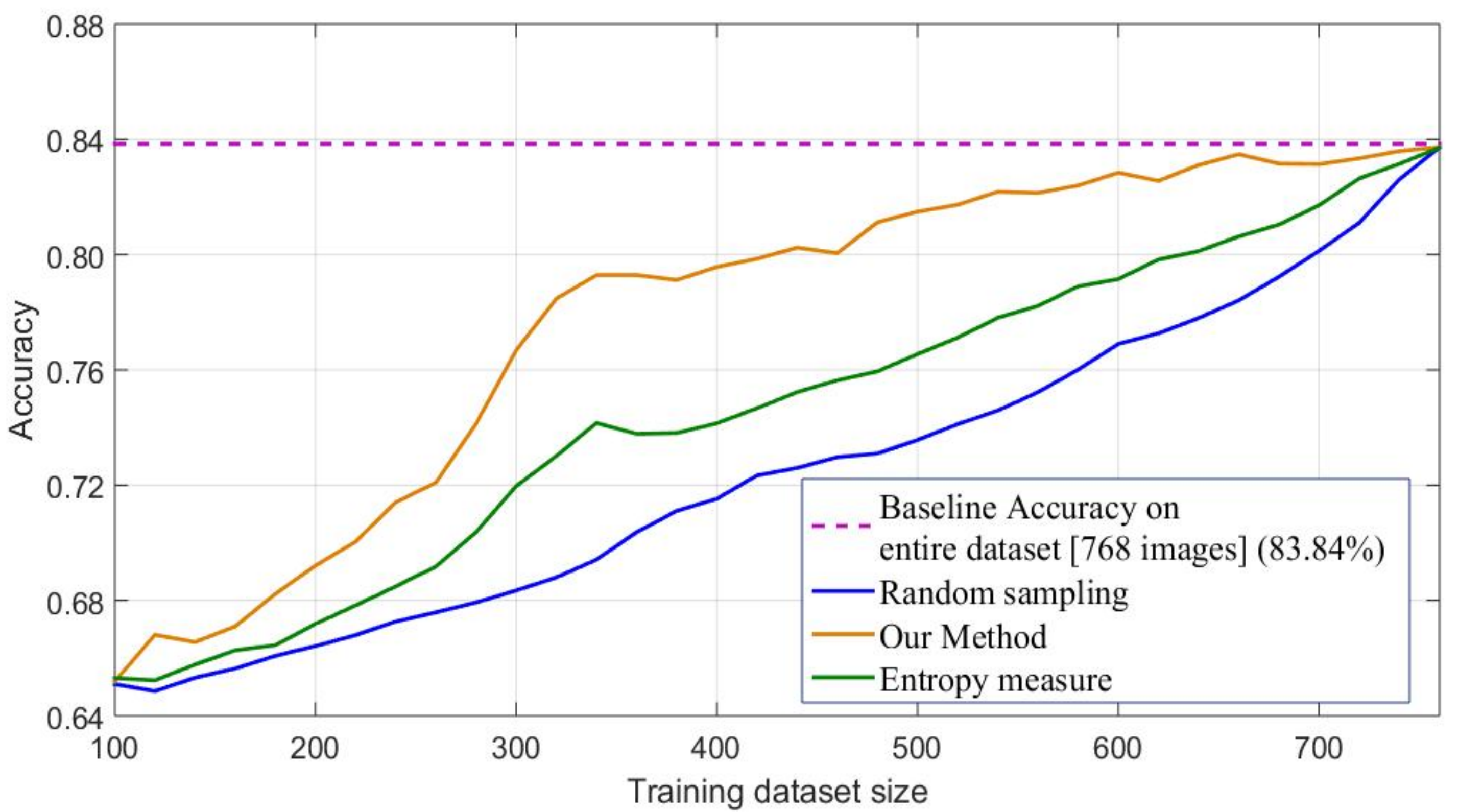}
\caption{\centering\textbf{Messidor: Test Accuracy.}}
\label{fig: messidor}
\end{subfigure}
\begin{subfigure}[t]{0.45\linewidth}
\centering
\includegraphics[width=\textwidth]{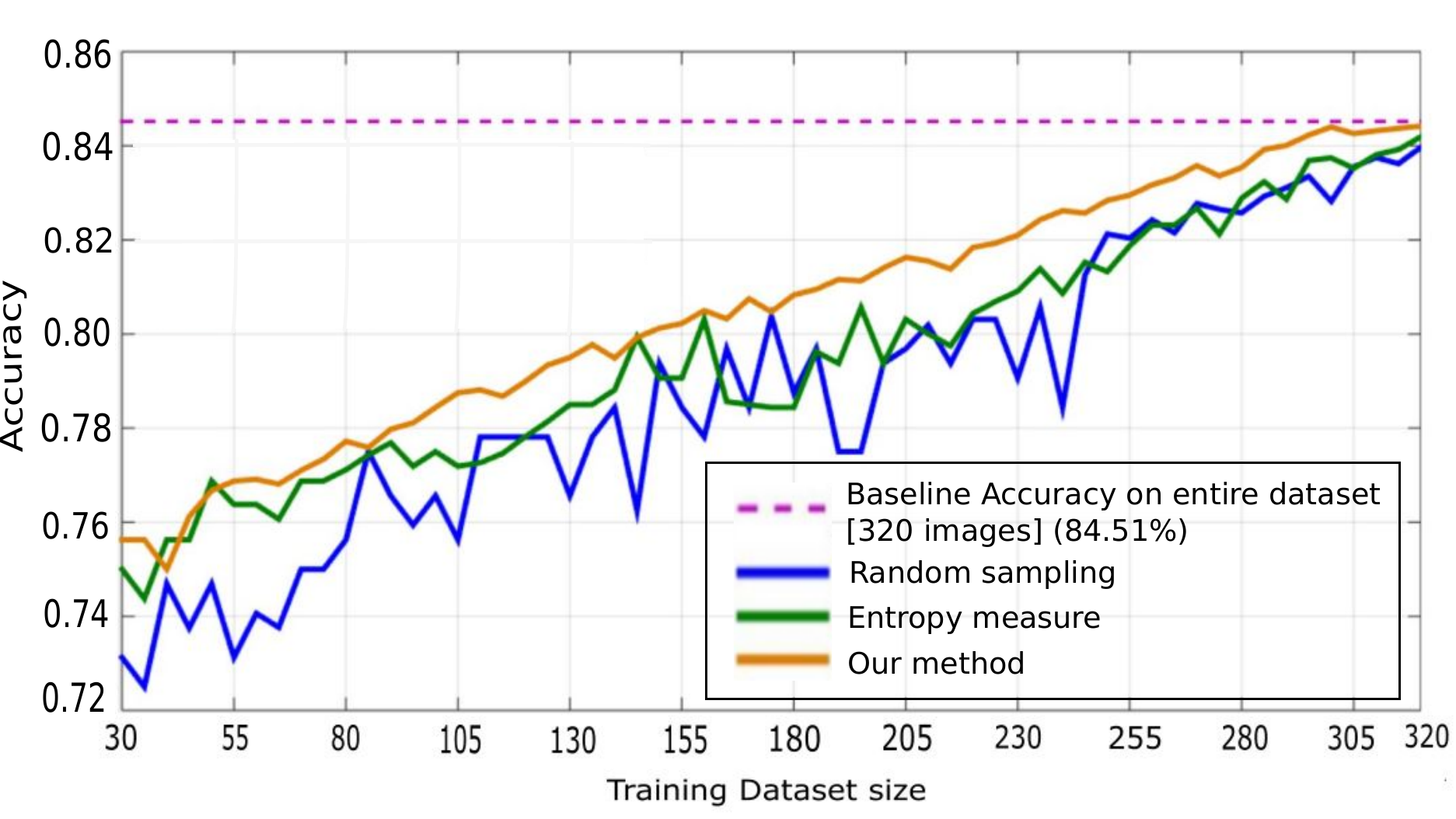}
\caption{\centering\textbf{Breast Cancer: Test Accuracy.}}
\label{fig: ICIAR}
\end{subfigure}
\begin{subfigure}[t]{0.45\linewidth}
\centering
\includegraphics[width=\textwidth]{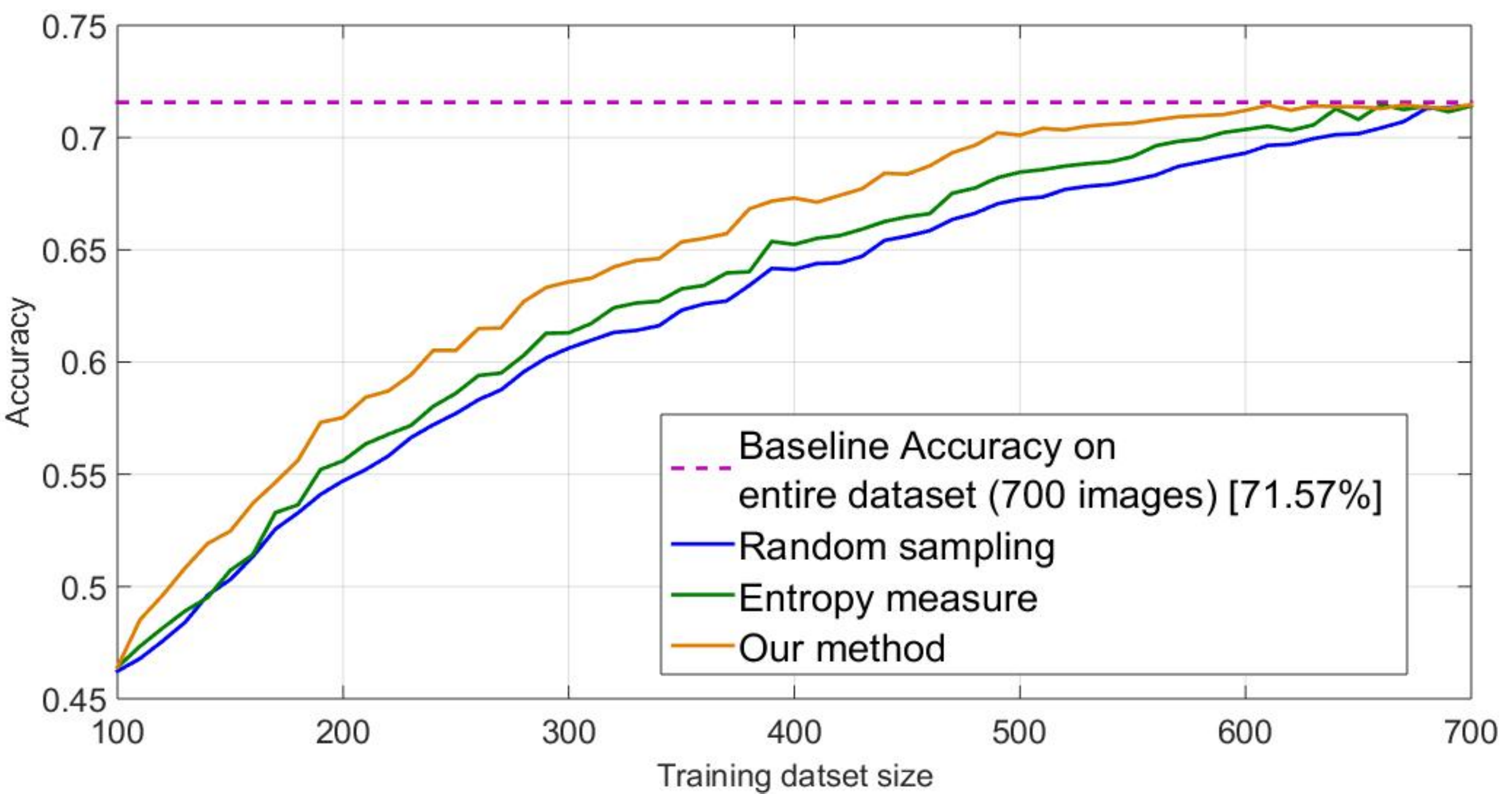}
\caption{\centering\textbf{Skin Cancer: Test Accuracy.}}
\label{fig: Skin_cancer_val}
\end{subfigure}
\begin{subfigure}[t]{0.45\linewidth}
\centering
\includegraphics[width=0.9\textwidth]{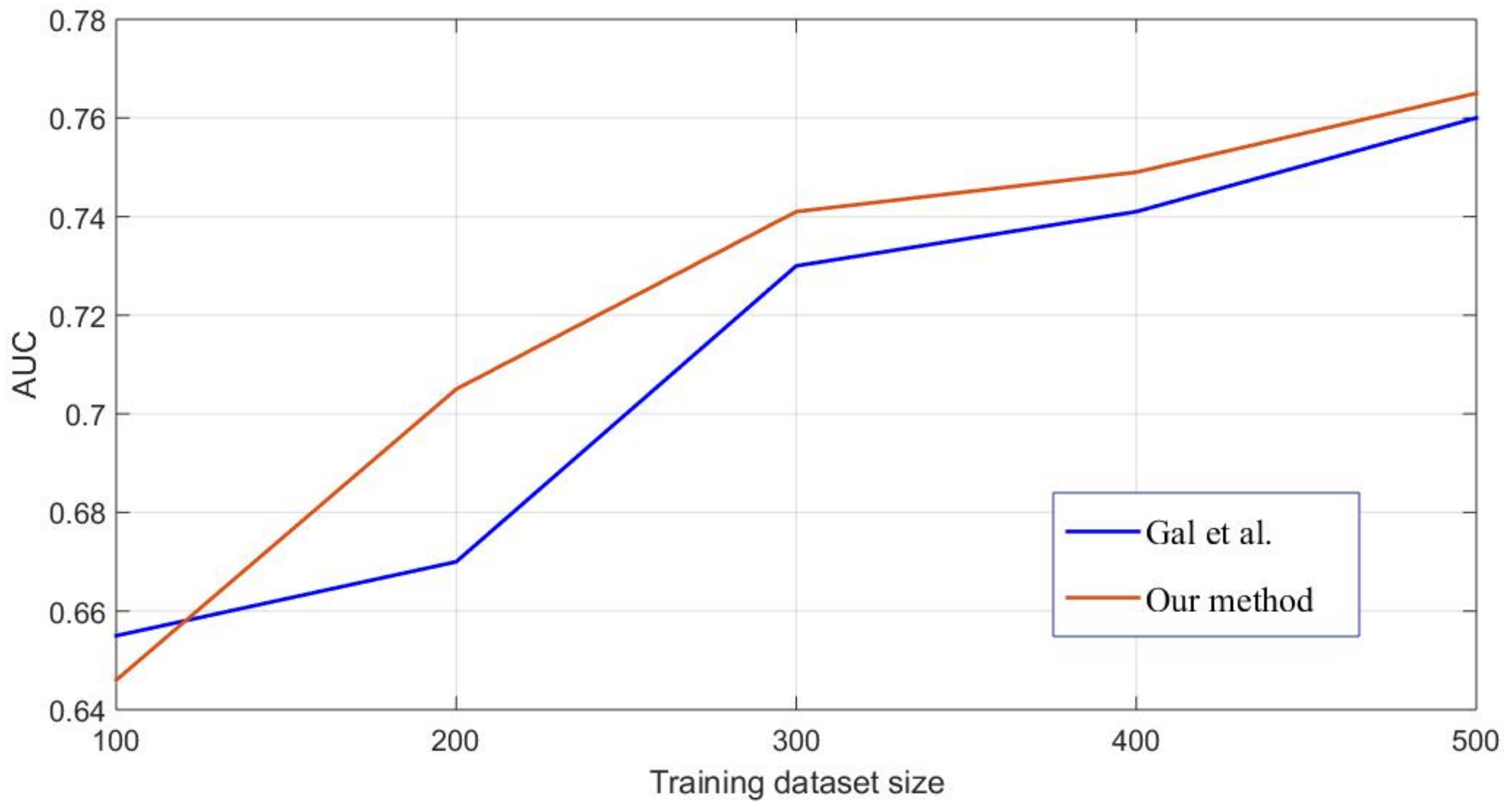}
\caption{\centering \textbf{Skin Cancer: Test AUC.}}
\label{fig: Skin_cancer_AUC}
\end{subfigure}
\caption{
MedAL outperforms competing methods by obtaining better performance with fewer labeled training images. Figures \textbf{(a)}, \textbf{(b)} and \textbf{(c)} show
that MedAL outperforms random sampling and uncertainty sampling (entropy measure) by consistently obtaining better test set accuracy as the size of the training set increases.
Figure \textbf{(d)} shows that MedAL outperforms a Deep Bayesian AL method \citep{gal2017deep} in terms of Area Under the ROC curve.
% \textbf{(a)} On Messidor, our method achieves $80\%$ accuracy with $200$ less labeled training images than uncertainty sampling. \\% Using our method, it is possible to obtain comparable results to a model trained on the full dataset of $768$ images using only $650$ labeled images (ie 84.6\% of dataset labeled). \\
% \textbf{(b)} On the Breast Cancer dataset, the improvement is less consistent.\\
% \textbf{(c)} On the unbalanced Skin Cancer dataset, our method achieves baseline accuracy with 90 images unlabeled.  \\
% \textbf{(d)} Our method slightly outperforms Gal, et al.'s Deep Bayesian AL method~\protect\citep{gal2017deep} on the Skin Cancer dataset.
 }
 \label{fig: medal_results}
\end{figure}

We compare the performance of our AL sampling method on the three datasets mentioned above to the performance of uncertainty sampling and random sampling. The datasets each present different learning challenges: Messidor as a balanced binary classification, the Breast Cancer dataset as a balanced multi-class classification, and the Skin Cancer dataset as an unbalanced binary classification.  After each AL iteration, the test accuracy of our model is evaluated.

Our method consistently outperforms both uncertainty and random sampling on all three datasets, as shown in Figure \ref{fig: medal_results}.  Random sampling chooses images uniformly at random, and therefore on balanced datasets, we expect accuracy to increase in a nearly linear fashion.  Any improvement over random sampling will choose more informative images earlier in the sampling process, resulting in a non linear curve.  MedAL shows large increases in accuracy with fewer images.  For instance, on Messidor, Figure \ref{fig: messidor} shows our method obtains $80\%$ accuracy with 425 images, whereas uncertainty sampling and random sampling require 625 and 700 images, respectively, to achieve the same $80\%$ accuracy. Moreover, our method obtains results comparable to the baseline accuracy using only 650 of 768 training images (84.6\%).   

As shown in Figure \ref{fig: ICIAR}, our method is also consistently better than competing methods in the Breast Cancer dataset, although the difference is not as striking as in Messidor.
Our approach reaches $82\%$ accuracy with 230 of 320 (76.9\%) images labeled, whereas uncertainty sampling and random sampling require 250 and 255 images, respectively, to achieve the same accuracy.

Finally, our approach also reaches $69\%$ accuracy on the Skin Cancer dataset with 460 images labeled, as shown in Figure \ref{fig: Skin_cancer_val}, whereas uncertainty sampling and random sampling require 570 and 640 labeled training images, respectively, to achieve the same results. Furthermore, our method achieves baseline accuracy of $71\%$ after being trained with 610 of 700 images (87\%).

MedAL's consistently superior performance on the Breast and Skin Cancer datasets show that the distance based sampling method is robust to different properties in the given unlabeled dataset.  In particular, we show robustness to balanced and imbalanced class distributions for binary and multi-class classification tasks.

\begin{figure}[t]
\centering
\includegraphics[width=\linewidth]{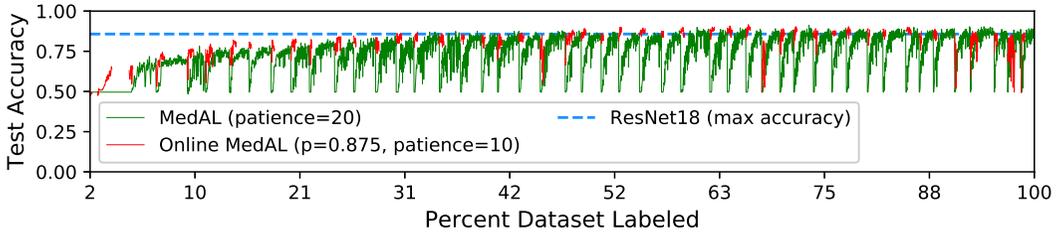}
\caption{\textbf{O-MedAL vs. MedAL Mechanics}  O-MedAL (red) preserves the labeling efficiency of MedAL (green) with less computation and higher accuracy, showing robustness to imbalance in the unlabeled data, \oracle.}
\label{fig: online_vs_MedAL}
\end{figure}

\begin{figure}[t]
\centering
\begin{subtable}{\textwidth}
\centering
\vspace{0.5cm}
\definecolor{b}{rgb}{.5,0,.7}
\definecolor{b2}{rgb}{.35,0,.55}
\definecolor{bl}{rgb}{.0,.0,.0}
\definecolor{g}{rgb}{0,.6,0}

\definecolor{LightGreen}{rgb}{.8,.9,.8}
\definecolor{LightPurple}{rgb}{.9,.7,1}
\definecolor{LightGray}{rgb}{.9,.9,.9}
\small
\begin{tabular}{llllll}
\toprule
                                  &        Experiment   &           Max Test Accuracy &   Percent Labeled &         Examples Processed      & Wall Time \\
\midrule
  \rowcolor{LightGreen}
\cellcolor{g}\textcolor{white}{+} &%OMedAL, p=0.875     &  \textbf{91.60\% (+5.88\%)} &           80.08\% &          128716 (+69.54\%)        & 1:09\\
                                    OMedAL p=0.875, patience=5 &  \textbf{92.02\% (+6.30\%)} &           90.62\% &         177658 (+134.01\%) & 1:29 \\
                                    % OMedAL p=0.875, patience=10 & \textbf{91.60\% (+5.88\%)} &           67.44\% &         195099 (+156.98\%) & 1:49 \\
  \rowcolor{LightGreen}
\cellcolor{g}\textcolor{white}{X} & MedAL (patience=20) &\underline{91.18\% (+5.46\%)}&           73.76\% &       1178079 (+1451.74\%)        & 10:06\\
  \rowcolor{LightPurple}
\cellcolor{b}\textcolor{white}{+} &%&OMedAL, p=0.875     &           85.71\% (+0.00\%) &\underline{40.04\%}&           31090 (-59.05\%)       & \underline{0:30}\\
                                    OMedAL p=0.875, patience=20 &  85.71\% (+0.00\%) &  \textbf{25.29\%} &           48335 (-36.33\%)         & 0:56\\

  \rowcolor{LightPurple}
  \cellcolor{b2}\textcolor{white}{X} & MedAL (patience=20) &           86.97\% (+1.26\%) &  \underline{33.72\%} &\underline{265106 (+249.19\%)}& 4:14 \\
  \rowcolor{LightGray}
\cellcolor{bl}\textcolor{white}{+}& OMedAL, p=0.125     &           85.71\% (+0.00\%) &           80.08\% &  \textbf{24296 (-67.82\%)}        & 0:54 \\
  % \rowcolor{LightGray}
% \cellcolor{bl}\textcolor{white}{X}& MedAL (patience=10) &           86.13\% (+0.42\%) &           90.62\% &\underline{1120482 (+1375.87\%)}   & 8:48 \\
% \cellcolor{b}\textcolor{white}{X} & MedAL (patience=20) &           86.97\% (+1.26\%) &  \underline{33.72\%} &         265106 (+249.19\%)     & 4:14 \\

                                  & OMedAL, p=0.125, patience=5    &85.71\% (+0.00\%) &           48.47\% &         51759 (-31.82\%)          & \underline{0:39}\\
                                  & ResNet18 Baseline   &           85.71\% (+0.00\%) &          100.00\% &            75920 (+0.00\%)        & \textbf{0:22} \\
\bottomrule
\end{tabular}

\caption{\centering\textbf{Experimental Results}}
\end{subtable}
\begin{subfigure}{0.49\linewidth}
\centering
\includegraphics[width=\textwidth]{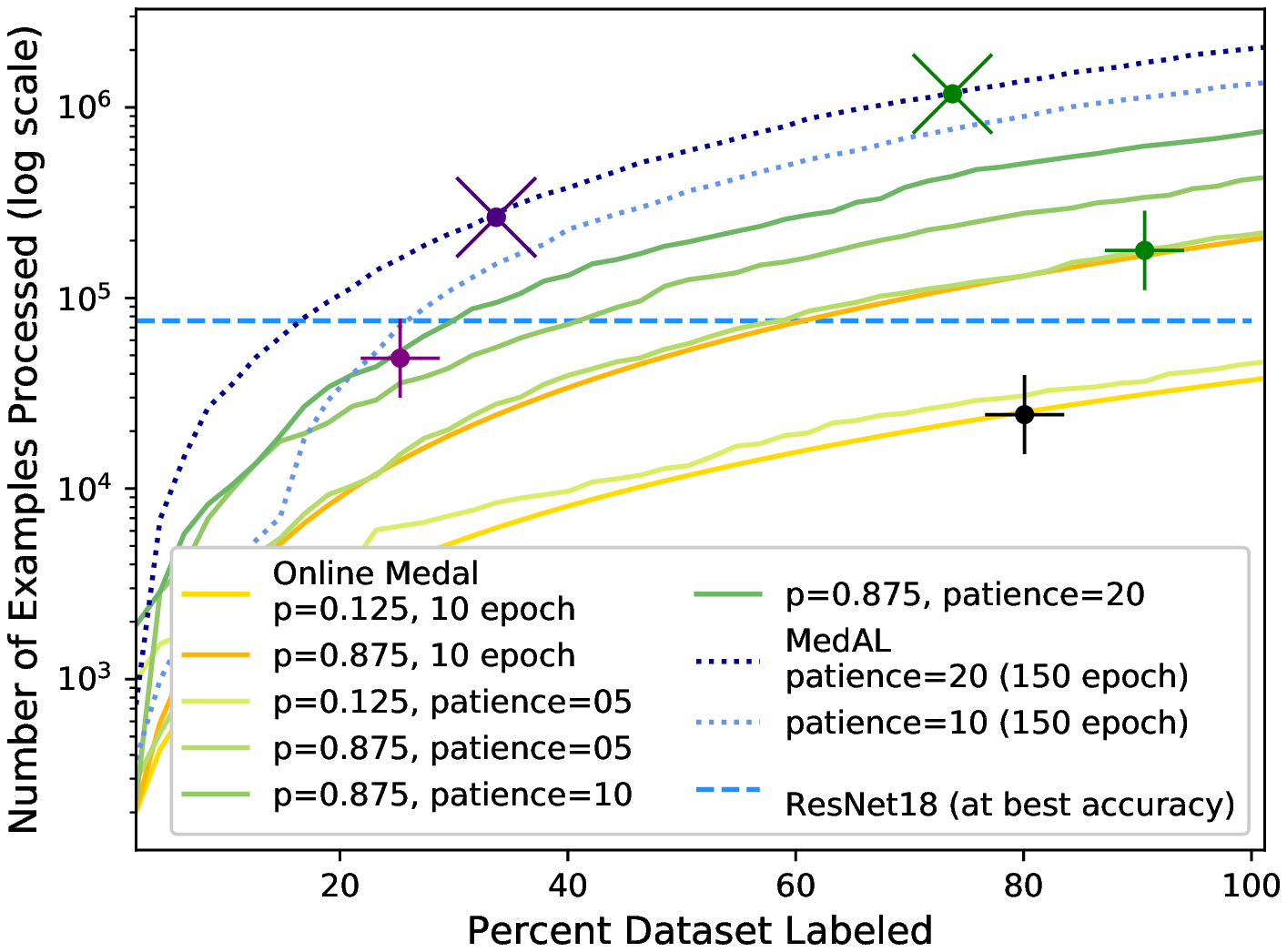}
\caption{\centering\textbf{O-MedAL Computational Efficiency}}
\label{fig:numimagesprocessed}
\end{subfigure}
\begin{subfigure}{0.49\linewidth}
\centering
\includegraphics[width=\textwidth]{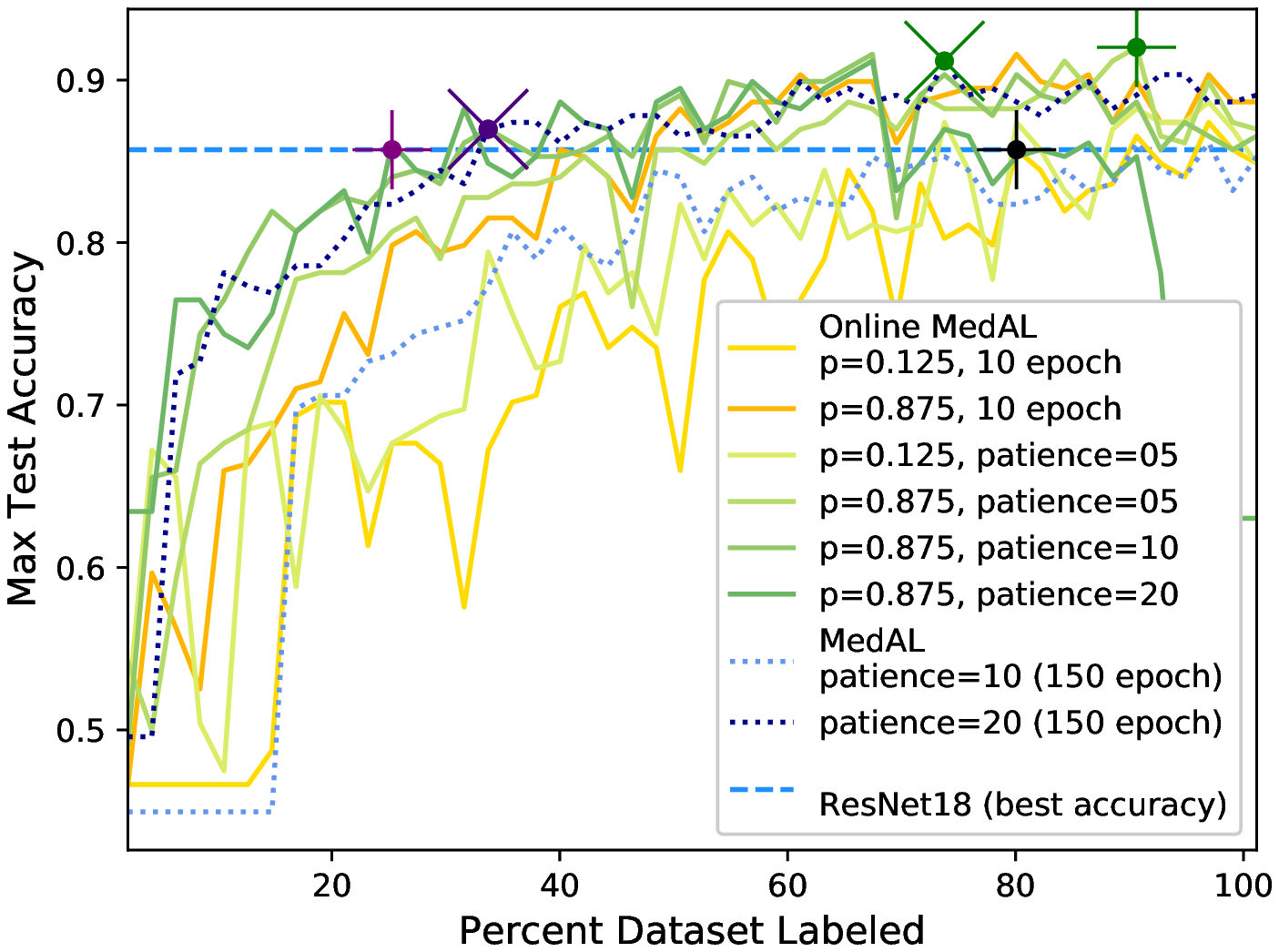}
\caption{\centering\textbf{O-MedAL Accuracy}}
\label{fig:topn_best}
\end{subfigure}
\caption{\textbf{O-MedAL Evaluation}.  Showing how computational efficiency (b) and accuracy (c) relate to labeling efficiency.
Marked points and table (a) highlight which MedAL and O-MedAL models perform best in each category.}
\label{fig: O-MedAL_evaluation}
\end{figure}

\subsection*{\sffamily \large Online Active Learning (O-MedAL) Evaluation}\label{sec:online_eval}

We evaluate O-MedAL by comparing it to MedAL and the ResNet18 baseline across three areas: test set accuracy, percent dataset labeled and computational efficiency. In particular, we conduct a study to determine whether previously labeled data is useful or necessary to train the model. Results show that the online approach significantly improves over the baseline in all three areas.

We train O-MedAL nine times independently, varying $p$, the percent of previously labeled images to include in any particular AL iteration.  The nine values we consider are $p \in \{0.0, 0.125, 0.25, 0.375, 0.5, 0.625, 0.75, 0.875, 1.0\}$.  These models all have a fixed 10 epochs per AL iteration.  We further evaluate the $p=0.125$ and $p=0.875$ models using patience of 5, 10 and 20, as these two values of $p$ were best performing.

Figure \ref{fig: online_vs_MedAL} shows the test accuracy (y-axis) as a function of percent dataset labeled (x-axis) for one O-MedAL (red line) and one MedAL model (green line).  Percent dataset labeled is directly proportional to AL iteration.  Within an AL iteration, we compute test accuracy after each epoch.  Therefore, the x axis can equivalently be labeled with a sequence of the form (AL iteration, epoch).  The dashed blue line is the baseline ResNet18 highest test accuracy (85.71\%).

The plot exposes key details of MedAL's learning process.  MedAL (green line) resets the weights of the model at each AL iteration and then trains for up to 150 epochs (subject to early stopping).  
The model starts at the same initialization (with 49.58\% test accuracy) and spends time redundantly re-learning the same features to regain the accuracy of the former AL iteration.

If MedAL converges before 100\% of dataset is labeled, it effectively re-trains the baseline model multiple times.  Assuming overfitting does not decrease test accuracy, we can expect the model's max accuracy to have exceeded the baseline accuracy due to random chance.  The results of Figure \ref{fig: online_vs_MedAL} confirm MedAL exceeded baseline accuracy.  An improvement in accuracy over MedAL's highest accuracy is thus a significant improvement over the baseline.  Secondly, we also see that the MedAL curve reaches baseline accuracy when approximately 61\% of the dataset is labeled, which is slightly better than our previous tests on MedAL with Inception V3.  Possible reasons for this improvement are (a) different dataset (the Messidor train set is larger, train and test sets are stratified across hospitals and the duplicate images removed), (b) not using ORB initialization, (c) a larger patience value (patience=20), (d) different implementation (PyTorch vs Keras).  %  We mention this possibility in the Discussion section. 

The red lines of Figure \ref{fig: online_vs_MedAL} represent our second most accurate O-MedAL model, which uses $p=0.875$, or $87.5\%$ of previously labeled data at each AL iteration, and has early stopping ($patience=10$).  Each small vertical red line represents the epochs of one AL iteration.  The lines are separated in order to visually align O-MedAL epochs to the MedAL epochs at each AL iteration.  We can see by the spacing between red lines that the model trains for fewer epochs than MedAL.  In contrast to the MedAL model, we also see O-MedAL's minimum accuracy at the start of each AL iteration is generally increasing, which suggests that the online method is able to learn incrementally without forgetting previous learnings.  % We plot this particular O-MedAL model because it is the only model that shows a strong drop in test accuracy after 69\% of the dataset is labeled.  The drop in performance suggests overfitting, since the online models train on the newly labeled images for each epoch of an AL iteration.  At later iterations, the images selected for labeling are most likely to be noise.  This result suggests that approximately 30\% of the Messidor dataset is entirely unhelpful to the binary classification of presence of Diabetic Retinopathy.  This effect is not seen in O-MedAL models with lower patience, likely because the models have not had a chance to overfit.  This effect is also not seen in MedAL models because each epoch evaluates the entire dataset, and we believe the noise is sufficiently drowned out by the informative images.  This aspect of the online models suggests O-MedAL could be useful for finding a subset of a dataset that is not useful to the learning task.  % Though not shown, we also found that increasing patience ... waiting on results TODO.  why show this model?  not sure of the argument I commented out.  TODO

The table and two subplots of Figure \ref{fig: O-MedAL_evaluation} show how O-MedAL exposes a three-way relationship between accuracy, percent dataset labeled and computational efficiency.  For MedAL and O-MedAL, respectively, the green markers highlight the most accurate models, the purple markers show the most data efficient models, and the black and dark purple markers show the most computationally efficient models.  Figure \ref{fig:numimagesprocessed} addresses computational efficiency while Figure \ref{fig:topn_best} addresses accuracy.  Both plots share an x axis of percent dataset labeled.  The dashed horizontal blue line represents the baseline ResNet18 model trained on the fully labeled dataset for 80 epochs.  The dotted blue lines are MedAL.  For clarity, we plot only the best performing O-MedAL models.

Figure \ref{fig:numimagesprocessed} plots the cumulative number of examples processed (on log scale) as a function of the percent dataset labeled.  The curve for each fixed epoch model can be computed before training using Equation \ref{eq:O-MedAL_num_processed}.  The curves for models using patience are not computed in advance of training, since early stopping leads to varying epochs per AL iteration.  MedAL and O-MedAL curves show exponential growth.  However, we show with the gray and purple markers that O-MedAL models reach baseline accuracy using less data and backpropagating fewer images than the baseline, suggesting O-MedAL can scale to larger datasets.  %Moreover, when $p \le 0.25$, O-MedAL is guaranteed to process less examples than the ResNet18 baseline run for 80 epochs.  If we use a baseline network that converges in more than 100 epochs, we expect O-MedAL's computational advantage will be more significant. This efficiency is useful because it offers flexibility in choice of $p$. 

To further examine the computationally efficiency of our online method, we measure total training time (wall time) of the shown models.  The right-most column of the table (5c) shows that the online method trains on Messidor between 4.5 to 6.8 times faster than MedAL, though no model is faster than the ResNet18 baseline.  This is primarily due to the fact that at each AL iteration, the algorithm evaluates the feature embedding for every labeled and unlabeled image, and secondarily due to the fact that we evaluated validation performance once per epoch.  O-MedAL models tested have approximately 500 to 2000 small epochs compared to ResNet18's 80 large epochs.  We evaluate validation performance each epoch, which is useful for the analysis in this paper, but leads to unnecessarily large wall time.  Indeed, validation performance in practice can be computed once per AL iteration.  The most interesting finding in the wall time analysis is that the O-MedAL model (p=0.875, patience=20) with the purple marker should be relatively computationally expensive model (as evidenced by Figure \ref{fig:numimagesprocessed}), yet it shows small wall time, processes fewer examples and achieves baseline accuracy with less labeled data than most of the models we tested.  This finding is counter-intuitive, yet offers useful practical advice when working with O-MedAL.  While one might assume a smaller model should train faster and process fewer examples, we found that the larger, more computationally expensive models can reach baseline accuracy just as efficiently using less labeled data.  This motivates generally choosing $p$ close to 1.

Figure \ref{fig:topn_best} plots test accuracy as a function of percent dataset labeled.  The $p=0.875$ O-MedAL models consistently show the highest accuracy over the AL iterations, which also means these models have the best data labeling efficiency.  As $p$ decreases, the model exchanges a decreased accuracy for fewer training examples backpropagated.  The low accuracy of the $p=0.125$ model is evidence that we need to include some amount of previously labeled data while training the model.  We also found that three fixed epoch O-MedAL models ($p \in \{0.375, 0.75, 0.875\}$) and all $p=0.875$ models with patience outperform the best MedAL model, which denotes a significant improvement over the baseline.  The fact that these models exceed the baseline accuracy is very important and suggests that OMedAL is able to find subsets of the training data that are more useful to the task than the full dataset.

Our results show that O-MedAL significantly improves the accuracy of its underlying ResNet18 model by as much as 6.3\%.  We also show that O-MedAL can reach the same performance as its baseline model using only 25\% of the training dataset.  Third, we show that O-MedAL is even able to backpropagate 67\% fewer examples than the ResNet18 model, with modest increase in wall time.

% \section{Discussion and Future Work}\label{sec:discussion}
\section*{\sffamily \Large DISCUSSION AND FUTURE WORK}\label{sec:discussion}

On each of the three datasets tested, MedAL achieves a higher overall accuracy and uses less labeled training data. The results confirm that informative examples are maximally dissimilar in feature space to previously trained examples and close to the decision boundary of the current model.

Our proposed active learning framework uses the trained baseline model to sample unlabeled examples.  As the model improves its ability to classify the data, it also naturally improves its ability to identify unlabeled examples worthy of labeling.  Since the method samples examples with maximum uncertainty and distance to the labeled examples, we can infer that its performance will continue to improve relative to other sampling techniques on larger unlabeled datasets.  We also show the method is robust to class imbalance in the unlabeled datasets by showing consistent improvement over competing methods on multi-class unbalanced data, multi-class balanced data, and binary balanced data via the Skin Cancer, Breast Cancer and Messidor datasets, respectively.
  Future work could attempt to quantify how the performance changes as a function of unlabeled dataset size and also measure how the class distribution of the unlabeled dataset changes.

Our online component to the active learning framework enhances the applicability of the framework to numerous scenarios.  The most pertinent is the medical setting where there is not enough labeled data to train a supervised deep model, but there exist human annotators interested in using a predictive model to enhance their diagnostic workflow.  For instance, in Diabetic Retinopathy, a physician must manually review high resolution retinal fundus images of Diabetic patients for microaneurysms that may be only a few pixels large.  O-MedAL enables the training of a deep model in such a setting, where it interactively asks the physician to annotate certain images most likely to improve the model's ability, and simultaneously returns a set of predictions on a set of images.  The online component enables near real-time interaction with a human annotator, where if we set the batch size to $M=1$, the model is as up-to-date as possible each time it queries an unlabeled example and generates a set of predictions.  Setting $M=1$ is desirable but intractable for many active learning methods, including MedAL without the online portion, where the annotator would need to wait hours between training events before labeling the next batch of images.  To the best of our knowledge, O-MedAL is the only existing framework enabling near real-time training of deep models on limited data in an active learning context.  As future work, we plan to incorporate O-MedAL into more real world medical applications with clinical annotators, like we have done in our EyeQual project paper \citep{eyequal}.

O-MedAL is presented as a solution for active learning, but the framework may have useful applications as a more generalized mechanism for training supervised deep learning models.  Further work should explore whether the 6\% improvement in accuracy of the baseline model generalizes to other architectures and datasets.  The ability to reach higher accuracy using less of the data suggests the framework can also be used to identify a subset of training examples that are not useful to a particular learning task.

% \section{Conclusions}\label{sec:conclusions}

\section*{\sffamily \Large CONCLUSIONS}\label{sec:conclusions}

In this paper we extend MedAL by introducing 
an online learning method. The online method retains all the benefits of MedAL while improving both its accuracy and computational efficiency.

We evaluate MedAL on three distinct medical image datasets: Messidor, the Breast Cancer dataset, and the Skin Cancer dataset. Our results show that MedAL is efficient, requiring less labeled examples than existing methods.  The method achieves comparable results to a model trained on the fully labeled dataset using only 650 of 768 images.  Additionally, MedAL attains 80\% accuracy using 425 images, corresponding to 32\% and 40\% reduction compared to uncertainty and random sampling methods, respectively.  Finally, we show that MedAL is robust, performing well on both binary and multi-class classification problems, and also on balanced and unbalanced datasets.  
MedAL shows consistently better performance than the competing methods on all three medical image datasets.

We compare O-MedAL to both MedAL and the underlying baseline deep network.  We show that O-MedAL is more accurate, improving test accuracy of its underlying ResNet18 deep network by 6.30\%.  We also show that O-MedAL achieves the baseline model's accuracy when only 25.29\% of the dataset is labeled.  Finally, O-MedAL is computationally efficient, reducing training wall time by 4.5 to 6 times over MedAL and processing up to 68\% fewer examples than the baseline ResNet18 with only 80\% of the dataset.  We have shown that O-MedAL is a mechanism for efficiently training deep supervised models in an active learning setting, and that it significantly improves performance of the underlying baseline model while using less data.

% \section*{acknowledgments}

% This work was supported in part by the European Regional Development Fund through the Operational Programme
% for Competitiveness and Internationalisation – COMPETE 2020
% Programme and in part by the National Funds through the
% Funda\c{c}\~{a}o para a Ci\^{e}ncia e a Tecnologia within under Project CMUPERI/TIC/0028/2014.

\section*{Conflict Of Interest}
The authors declare that they have no conflicts of interest for this work.

\printendnotes

% Submissions are not required to reflect the precise reference formatting of the journal (use of italics, bold etc.), however it is important that all key elements of each reference are included.
\bibliography{References}

% \graphicalabstract{Figures/fig1-MedAL_online.eps}{
  % We present a novel online active learning framework for efficiently training deep supervised models in an active learning setting.  O-MedAL significantly enhances its underlying baseline model in our experiments by improving the model's underlying deep network accuracy by 6.30\%, using only 25\% of the labeled dataset to achieve baseline accuracy, reducing backpropagated images during training by as much as 67\%, and demonstrating robustness to class imbalance in binary and multi-class tasks.
% }

\end{document}